%% file: main.tex
\documentclass[10pt,a4paper]{article}

\usepackage{url}
\usepackage{color}
\usepackage{graphicx}

\usepackage{authblk}

\title{Anticipation in collaborative music performance\\
  using fuzzy systems: a case study}

\author[1]{Oscar Th\"{o}rn}
\author[2]{Peter F\"{o}gel}
\author[2]{Peter Knudsen}
\author[3]{\\ Luis de Miranda}
\author[1]{Alessandro Saffiotti}

\affil[1]{School of Science and Technology
\par \"{O}rebro University, 70182 \"{O}rebro, Sweden
\par \protect\url{oscar.thorn@telia.com}, \protect\url{alessandro.saffiotti@oru.se}}

\affil[2]{School of Music, Theatre and Art
\par \"{O}rebro University, 70182 \"{O}rebro, Sweden
\par \protect\url{peter.fogel@oru.se}, \protect\url{peter.knudsen@oru.se}}

\affil[3]{School of Humanities, Education and Social Sciences
\par \"{O}rebro University, 70182 \"{O}rebro, Sweden
\par \protect\url{luis.de-miranda@oru.se}}

\date{\today}

\begin{document}

\maketitle

\begin{abstract}
  In order to collaborate and co-create with humans, an AI system must
  be capable of both reactive and anticipatory behavior.  We present a
  case study of such a system in the domain of musical improvisation.
  We consider a duo consisting of a human pianist accompained by an
  off-the-shelf virtual drummer, and we design an AI system to control
  the perfomance parameters of the drummer (e.g., patterns, intensity,
  or complexity) as a function of what the human pianist is playing.
  The AI system utilizes a model elicited from the musicians and encoded
  through fuzzy logic.  This paper outlines the methodology, design, and
  development process of this system.  An evaluation in public concerts
  is upcoming.  This case study is seen as a step in the broader
  investigation of anticipation and creative processes in mixed
  human-robot, or ``anthrobotic'' systems.
\end{abstract}

\newpage
\input{intro}
\input{method}

\input{system}

\input{develop}

\input{conclusions}

\section*{Acknowledgements}

This work was partially supported by a grant from Vetenskapsr\aa{}det.
Additional support was provided by \"{O}rebro University through its
Internationalization program as well as its 20-year jubileum program.
The collaboration that led to this work is framed within the CREA
(Cross-disciplinary Research on Effectual Anticipation) research
network~\cite{crea.web}, which is part of \"{O}rebro University's larger
effort to promote multidisciplinary research in AI.

\bibliographystyle{plain}
\bibliography{music}

\end{document}

%% file: intro.tex
\section{Introduction}
\label{intro.sec}

The creation and performance of music has inspired AI researchers since
the very early times of artificial
intelligence~\cite{HillerIsaacson.jaes1958,SimonSumner.mmm1968,Minsky.mmb1982},
and there is today a rich literature of computational approaches to
music~\cite{MorEtAl.acme2019}, including AI systems for music
composition~\cite{Cope.book2005} and improvisation~\cite{Biles.leo2003}.
As pointed out by Thom~\cite{Thom.um2003}, however, these systems rarely
focus on the spontanous interaction between the human and the artificial
musicians.  We claim that such interaction demands a combination of
reactivity and anticipation, that is, the ability to act now based on a
predictive model of the companion player~\cite{Rosen.book2012}.

This paper reports our initial steps in the generation of collaborative
human-machine music performance, as a special case of the more general
problem of anticipation and creative processes in mixed human-robot, or
\emph{anthrobotic} systems~\cite{DeMirandaEtAl.rp2016}.  We consider a
simple case study of a duo consisting of a human pianist accompained by
an off-the-shelf virtual drummer, and we design an AI system to control
the key perfomance parameters of the virtual drummer (patterns,
intensity, complexity, fills, and so on) as a function of what the human
pianist is playing.  The AI system is knowledge-based: it relies on an
internal model represented by means of fuzzy logic.  This model encodes
the expertise of musicians about joint music performance, elicited
through a process of user-centered design.  Musicians have provided
heuristic rules using vague linguistic terms, which are suitably encoded
using the tools of fuzzy logic.  Note that, while rule-based systems
have been often used for music composition and
improvisation~\cite{TatarPasquier.jnmr2019}, the use of fuzzy logic in
this field is much less explored~\cite{LiuChuanKang.etci2017}.

The knowledge model in our system includes both reactive responses to
the current parameters of the pianist performance, like intensity or
rhythmic complexity, and anticipatory responses to forecasted events,
like the coming climax of a crescendo.  We take input from the piano
player in real time via a MIDI interface, extract meaningful musical
parameters like intensity, rythmic complexity, or crescendo.  The model
is used to generate as output musical parameters that control the
execution of a Strike\,2 virtual drummer.

The rest of this paper is organized as follows.  The next section
summarizes the general methodology adopted in our work, and the specific
case study reported in this paper.  Section~\ref{system.sec} gives some
technicalities on the implmented system, while Section~\ref{develop.sec}
focuses on the development and testing process.
Section~\ref{conclusions.sec} discusses the results and concludes.


%% file: method.tex
\section{Methodology}
\label{method.sec}

As mentioned above, in this paper we are not interested in fully
automated music composition.  Rather, we are interested in the
collaborative execution between a human musician and a robotic
performer.  We assume that the robotic performer is capable of
autonomous artistic execution, and that the modalities of this execution
are controlled by a fixed number of parameters.  This paper addresses
the problem of controlling the parameter of execution in order to obtain
a harmonious joint performance. 

\begin{figure}[t]
  \centering
  \includegraphics[width=120mm]{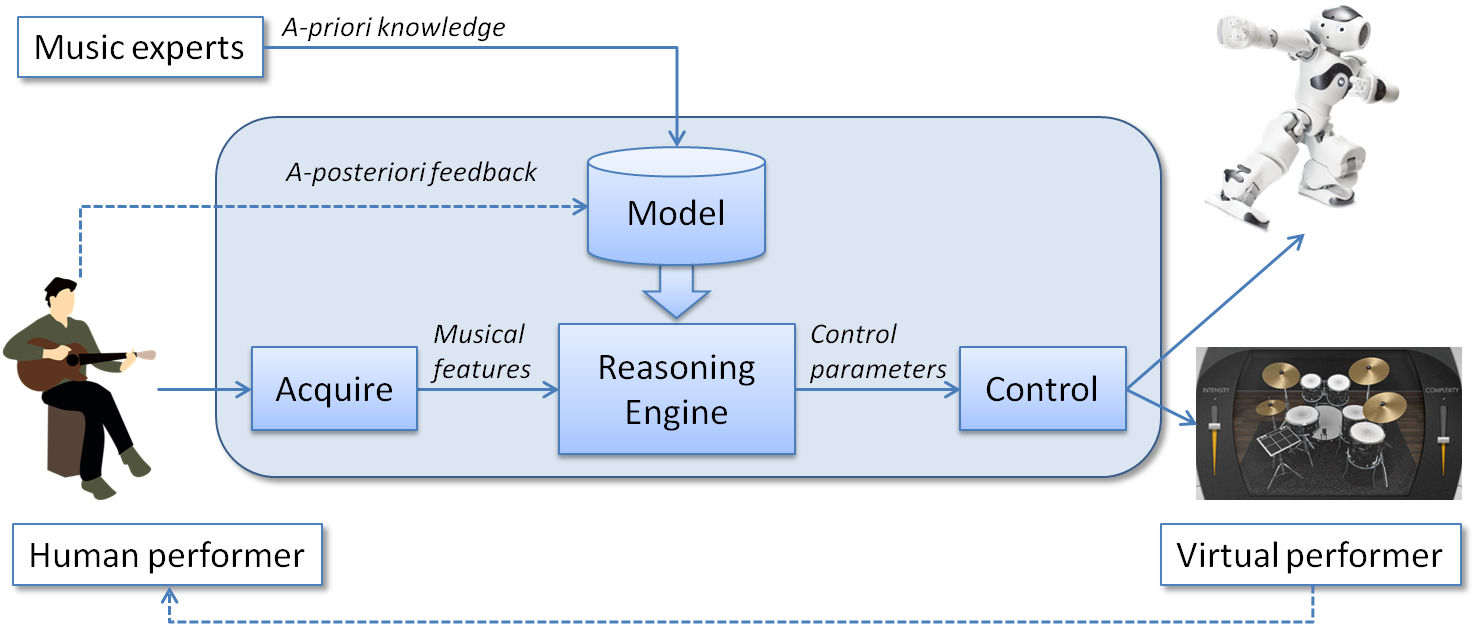}
  \caption{The proposed methodology to control a virtual music partner}
  \label{concept.fig}
\end{figure}

Figure~\ref{concept.fig} illustrates the concept used in this paper.  A
human musician plays freely, and an AI system controls the parameters of
a robotic performer accordingly.  We use ``robotic'' here in a broad
sense to mean any agent that generates physical actions.  This could be,
for instance, a dancing robot, a virtual drummer, or a sound processing
agent that spatializes in the hall the music produced by the human
musician.  The dependencies of the parameters of the robotic performer
on the features of the human's execution are encoded into an explicit
model.  This is initially obtained by eliciting knowledge from the
musicians: later on, it can be refined and adapted through machine
learning techniques.

In the specific case addressed in this paper, the musician is a jazz
pianist and the artificial artist is a Strike~2 virtual drummer.%
\footnote{AIR Music Technology, https://www.airmusictech.com/product/strike-2/.}
The drums parameters controlled by the AI system include patterns,
intensity, complexity, fills, instruments used, and enter or exit
sequences.  Acquisition is done through a MIDI interface and a feature
extraction algorithm; controlling the drums parameters is done through
MIDI.  Upon discussions with the musicians, it turns out that part of
the knowledge of how the drummer's parameters depend on the pianist's
play is conscious, and the musicians can easily expressed it in terms of
approximate rules using vague linguistic terms, like:

\begin{quote}
  \textbf{If} the rhythmic complexity on the lower part of the keyboard
  is high,
  \\ 
  \textbf{then} the rhythmic complexity of the drums should increase.
\end{quote}

Fuzzy logic offer suitable tools to encode this type of knowledge, and
therefore we use it in our system.  The next section outlines the
technical choices made in our design.


%% file: system.tex
\section{System Design}
\label{system.sec}



The core of our system is a multiple-input multiple-output Fuzzy
Inference System (FIS)~\cite{Driankov.book2001}, which implements the
``Reasoning engine'' block in Figure~\ref{concept.fig}.  The system runs
at a fixed clock cycle, and it therefore resembles the structure of a
classical fuzzy controller.  It takes as input a set of music parameters
extracted in real time that describe the human execution, and it
produces as output a set of control parameters for the virtual drummer.
Differently from most conventional fuzzy controllers, the rules'
conditions are not simply conjunctions of positive literals, but general
formulas in (fuzzy) propositional logic.  This gives us greater
expressive power in representing the musician's knowledge.  Because of
this, the system has been implemented from scratch rather than relying
on existing toolboxes.


\subsection{Input features}

The interface between the software and the live or prerecorded piano is
implemented using the Python MIDO library. MIDI uses ports as interfaces
between producers and consumers of MIDI messages and the system will
continually poll the input port for MIDI messages. Form this the system
extracts features both explicit and implicit. The features extracted
are:

\begin{itemize}

\item \textit{Velocity.} The velocity is extracted from the velocity
  field of each received MIDI messages and the sum of velocities
  extracted at each clock cycle is given as input to the FIS.

\item \textit{Rhythmic density.} The rhythmic density is extracted for
  low, high and full range of notes by subdividing a musical bar into a
  number of slots and determining in which proportion of those slots a
  note has been played during the last bar.

\item \textit{Status of pedal.}  In MIDI the changing the pedal of a
  piano generates a control change message, our system keeps track of
  the current state of the pedal and changes it when receiving such a
  message.

\item \textit{Time since last piano note.} Is calculated at each
  iteration of computation from the recorded time of the last note.

\item \textit{Current musical bar, beat and position in 32 bar
  sequence.} Current musical bar and current beat is calculated by
  recording the time each note arrives and the time the first note
  (which should arrive on the first beat) arrived respectively.

\item \textit{Velocity averages.} A window of length t is created where
  each arriving velocity is recorded. At each computational cycle the
  window is dived into two part and an average is calculated for each
  part. This gives a newer and an older average

\end{itemize}

\subsection{Filtering the past}

Some of the knowledge expressed by the musicians implicitly refers to a
temporal aspect, e.g., considering the ``average'' intensity rather than
the instantaneous one, or considering that the intensity is
``increasing''.  These aspects could be captured in the feature
extraction part by adding ad-hoc temporal filters.  We opted instead for
using a second FIS to extract relevant temporal features.  This is a
recurrent fuzzy system~\cite{AdamyKempf.fss2003} that takes as input the
current features at time $t$ plus its own output at time $t-1$.  This
solution allows us to better capture the specific knowledge of the
musician, e.g., on what counts as a ``sudden drop in intensity'', in a
way that is more explicit and easier to modify. This FIS takes the
following variables as input:

\begin{itemize}

\item \textit{Velocity Difference.} The difference between the
  instantaneous velocity and the previous average velocity. U: (-127,
  127) Member functions shape: Triangular. Terms: 'Neg-Max' (-127, -127,
  -95.25), 'Neg-High' (-127, -95.25, -63.5), 'Neg-Middle' (-95.25,
  -63.5, -31.75), 'Neg-Low' (-63.5, -31.75, 0.0), 'None' (-31.75, 0.0,
  31.75), 'Low' (0.0, 31.75, 63.5), 'Middle' (31.75, 63.5, 95.25),
  'High' (63.5, 95.25, 127), 'Max' (95.25, 127.0, 127).

\item \textit{Density Difference.} The difference between the current
  rhythmic density and the previous rhythmic density, calculated for the
  high, low and full registers of notes. U: (-127, 127) Member functions
  shape: Triangular. Terms: 'Neg-Max' (-127, -127, -95.25), 'Neg-High'
  (-127, -95.25, -63.5), 'Neg-Middle' (-95.25, -63.5, -31.75), 'Neg-Low'
  (-63.5, -31.75, 0.0), 'None' (-31.75, 0.0, 31.75), 'Low' (0.0, 31.75,
  63.5), 'Middle' (31.75, 63.5, 95.25), 'High' (63.5, 95.25, 127), 'Max'
  (95.25, 127.0, 127).

\item \textit{Newer and older velocity average.} Referees to the two
  averages described in Input Features. U: (0, 127). Member functions
  shape: Triangular. Terms: 'Low' (0, 0, 25.4) 'Mid-Low' (0, 25.4,
  50.8), 'Middle' (25.4, 50.8, 76.2), 'Mid-High' (50.8, 76.2, 101.6),
  'High' (76.2, 101.6, 127), 'Max' (101.6, 127, 127).

\item \textit{Complexity.} The current complexity level of the piano as
  inferred by the control FIS. U: (0, 127). Member functions shape:
  Triangular. Terms: 'Low' (0, 0, 25.4) 'Mid-Low' (0, 25.4, 50.8),
  'Middle' (25.4, 50.8, 76.2), 'Mid-High' (50.8, 76.2, 101.6), 'High'
  (76.2, 101.6, 127), 'Max' (101.6, 127, 127).

\item \textit{Intensity.} The current intensity level of the piano as
  inferred by the control FIS. U: (0, 127). Member functions shape:
  Triangular. Terms: 'Low' (0, 0, 25.4) 'Mid-Low' (0, 25.4, 50.8),
  'Middle' (25.4, 50.8, 76.2), 'Mid-High' (50.8, 76.2, 101.6), 'High'
  (76.2, 101.6, 127), 'Max' (101.6, 127, 127).

\item \textit{Intensity Slope.} Considers if the intensity is increasing
  or decreasing (i.e. the slope is positive or negative). The slope is
  calculated by linear regression over k measured points of intensity.
  U: (-\(\infty\), \(\infty\)). Member functions shape:
  Trapezoid. Terms: 'Increasing' (1, 10, \(\infty\), \(\infty\)),
  'Decreasing'  (-1, -10, \(\infty\), \(\infty\)).

\item \textit{Complexity Slope.} Considers if the complexity is
  increasing or decreasing (i.e. the slope is positive or negative). The
  slope is calculated by linear regression over k measured points of
  complexity.  U: (-\(\infty\), \(\infty\)). Member functions shape:
  Trapezoid. Terms: 'Increasing' (1, 10, \(\infty\), \(\infty\)),
  'Decreasing'  (-1, -10, \(\infty\), \(\infty\)).

\end{itemize}

These antecedent variables are then used to infer the consequent
variables that are given as input to the control FIS.

\begin{itemize}

\item \textit{Change Velocity.} Depends on Velocity Difference and gives
  a scalar value as output that is used to calculate the new
  average. This consequent is inferred multiple times with different
  universe (-0.3 to 0.3 and -0.1 to 0.1) which makes it possible to
  calculate averages that change at different paces. U: (-0.3,
  0.3). Member functions shape: Triangular. Terms: 'Max-Down' (-0.3,
  -0.3, -0.2249), 'High-Down' (-0.3, -0.2249, -0.149), 'Middle-Down'
  (-0.2249, -0.1499, -0.0749), 'Low-Down' (-0.1499, -0.0749, 0), 'None'
  (-0.0749, 0, 0.0750), 'Low-Up' (0, 0.0750, 0.1500), 'Middle-Up'
  (0.07500, 0.1500, 0.2250), 'High-Up' (0.1500, 0.2250, 0.3), 'Max-Up'
  (0.2250, 0.3, 0.3).

\item \textit{Change Density.} Depends on Complexity Difference and
  gives a scalar value as output that is used to calculate the new
  average. This consequent is inferred multiple times for different
  parts of the register of notes. U: (-0.5, 0.5). Member functions
  shape: Triangular. Terms: 'Max-Down' (-0.5, -0.5, -0.375), 'High-Down'
  (-0.5, -0.375, -0.25), 'Middle-Down' (-0.375, -0.25, -0.125),
  'Low-Down' (-0.25, -0.125, 0.0), 'None' (-0.125, 0.0, 0.125), 'Low-Up'
  (0.0, 0.125, 0.25), 'Middle-Up' (0.125, 0.25, 0.375), 'High-Up' (0.25,
  0.375, 0.5), 'Max-Up' (0.375, 0.5, 0.5).

\item \textit{Sudden shift.} Depends on newer and older average,
  determines if there has been a sudden shift in the intensity of the
  piano playing, the time of a sudden shift is also recorded. U: (-1,
  1). Member functions shape: Triangular. Terms: 'Up' (1, 1, 0),
  'None'(1, 0, -1), 'Down' (0, -1, -1).

\item \textit{Hype.} Depends on current Intensity and Complexity as well
  as their slopes. Anticipates the arrival of a crescendo. U: (0,
  1). Member functions shape: Triangular. Terms: Coming' (0, 1, 1).

\end{itemize}

\subsection{Anticipating the future}

Another important element that we want to address is musical
anticipation.  We have encoded a simple predictive model in the above
temporal FIS to infer a coming climax or anti-climax from a change in
intensity and complexity.  The main FIS includes anticipatory rules that
react to these forecasted features, e.g., anticipate a climax by
starting a drums fill-in; or anticipate an anti-climax by muting the
kick first, and then the snare once the change occurs.

\subsection{Output parameters}

The virtual drummer is controlled by the software Strike 2 by Air Music
Technology. Strike allows us to control the behaviour and settings of
the drummer by the use of MIDI messages outputted from our
software. Currently our software controls the intensity and complexity
of the drummer as well as starting, stopping and changing the pattern
(e.g., verse, bridge, chorus, fills, intros and outros) of the drummer
and muting of individual parts of the kit.

\subsection{Fuzzy inference}

Our FIS is based on the usual fuzzify-inference-defuzzify pipeline. But
we use two fuzzy inference systems, one to infer temporal aspects of the
piano playing and one to infer the control parameters passed to the
drummer.

\subsection{Inferring the drum parameters}

The control FIS infers the commands that controls the behaviour of the drum-machine based on both the output of the temporal FIS and features extracted from the piano.

\begin{itemize}

\item \textit{Time since last note.} Refers to the time since the last
  piano note. U:(0, \(\infty\)). Member functions shape:
  Triangular. Term: Short (0, 4, 6)

\item \textit{Current Mode.} Refers to the current inferred mode of the
  piano. U:(0, 1). Member functions shape: Singular. Terms: 'Stop' (0),
  'Play' (1).

\item \textit{Historic Mode.} Refers to the previous inferred mode of
  the piano. U:(0, 1). Member functions shape: Singular. Terms: 'Stop'
  (0), 'Play' (1).

\item \textit{Historic Pattern} Refers to the current state of the
  drums. U:(0,15). Member functions shape: Singular. Terms: 'Intro To
  Chorus 1' (0), 'Fill To Chorus 1' (1), 'Fill To Chorus 2' (2), 'Fill
  To Chorus 3' (3), 'Fill To Chorus 4' (4), 'Fill To Chorus 5' (5),
  'Fill To Chorus 6' (6), 'Fill To Chorus 7' (7), 'Chorus 1' (8),
  'Outro' (10), 'None' (11).

\item \textit{Historic Mute.} Refers to which, if any, of the parts of
  the kit has been muted. U:(0, 2). Member functions shape:
  Singular. Terms: 'None' (0), 'Kick' (1), 'Kick and Snare' (2).

\item \textit{Intensity.} Refers the inferred intensity. This is
  determined first by the FIS and given as input to another part of the
  FIS in the same cycle. U:(0, 127). Member functions shape:
  Triangular. Terms: 'Low' (0, 0, 25.4) 'Mid-Low' (0, 25.4, 50.8),
  'Middle' (25.4, 50.8, 76.2), 'Mid-High' (50.8, 76.2, 101.6), 'High'
  (76.2, 101.6, 127), 'Max' (101.6, 127, 127).

\item \textit{Complexity.} Refers the inferred complexity. This is
  determined first by the FIS and given as input to another part of the
  FIS in the same cycle. U:(0, 127). Member functions shape:
  Triangular. Terms: 'Low' (0, 0, 25.4) 'Mid-Low' (0, 25.4, 50.8),
  'Middle' (25.4, 50.8, 76.2), 'Mid-High' (50.8, 76.2, 101.6), 'High'
  (76.2, 101.6, 127), 'Max' (101.6, 127, 127).

\item \textit{Bar.} Refers to the current position in a repeating 32 bar
  sequence (i.e. modulo 32). The member functions depend on the length
  of a bar, which in turn depend on the BPM and the number of notes per
  bar. Generally for the beginning and end of a bar respectively, where
  T is the time per bar and K is the bar number:

\[((K-1) * T - \epsilon, (K-1) * T, (K-1) * T + 0.009, (K-1) * T + 0.009 + \epsilon)\]

\[((K+1) * T -(T/8) - \epsilon, (K+1) * T -(T/8), (K+1) * T - 0.1, (K+1) * T\]

  U:(0, 32 * T). Member functions shape: Trapetziod. Terms: '8th',
  '16th', '24th', '32th', 'End 4th', 'End 12th', 'End 20th', 'End 28th'.

\item \textit{Change Velocity.} Refers to if the velocity has been
  increasing or decreasing over the past 8 bars. U:(0, 1). Member
  functions shape: Singular. Terms: 'Down' (0), 'Up' (1).

\item \textit{Hype.} Refers to the output variable given by the temporal
  FIS. U:(0, 1). Member functions shape: Triangular. Terms: 'Coming' (0,
  1, 1).

\item \textit{Time in bar.} Refers to the current position within a
  single bar. U:(0, T). Only a single term exist 'Last Quarter' which
  has the general membership function of the trapezoid shape:

\[T - (T/4) - \epsilon, T - (T/4), T, T + \epsilon \]

\item \textit{Average Velocity.} Refers to the average velocity inferred
  by the temporal filter FIS. U: (0, 127). Member functions shape:
  Triangular. Terms: 'Low' (0, 0, 25.4) 'Mid-Low' (0, 25.4, 50.8),
  'Middle' (25.4, 50.8, 76.2), 'Mid-High' (50.8, 76.2, 101.6), 'High'
  (76.2, 101.6, 127), 'Max' (101.6, 127, 127).

\item \textit{Intensity shift.} Refers to sudden shifts in intensity
  inferred by the temporal filter FIS. U:(-1, 1) . Member function
  shape: Triangular. Terms: 'Down' (-1, -1, 0), 'Up' (0, 1, 1).

\item \textit{Time since sudden shift up/down}. Refers to the amount of
  time passed since a sudden shift up and down respectively. U:(0,
  \(\infty\)). Member functions shape: Trapezoid. Terms: 'Short' (0, 0,
  0.5, 3.5)

\item \textit{Full range density.} Refers to the full range density
  extracted from the piano input. U:(0, 127). Member functions shape:
  Triangular. Terms: 'Low' (0, 0, 25.4) 'Mid-Low' (0, 25.4, 50.8),
  'Middle' (25.4, 50.8, 76.2), 'Mid-High' (50.8, 76.2, 101.6), 'High'
  (76.2, 101.6, 127), 'Max' (101.6, 127, 127).

\item \textit{Low/High Range Density.} Refers to the low/high range
  density extracted from the piano input. U:(0, 127). Member functions
  shape: Triangular. Terms: 'Low' (0, 0, 42.3), 'Middle' (0, 42.3,
  84.6), 'High' (42.3, 84.6, 127), 'Max' (84.6, 127, 127).

\item \textit{Pedal status.} Refers to the status of the pedal. U:(0,
  1). Member functions shape: Singular. Terms: 'Down' (0), 'Up' (1).

\item \textit{Time since pedal.} Refers to the amount of time since the
  sustain pedal was pressed down. U:(0, \(\infty\)). Membership
  functions shape: Triangular. Terms: 'Very Short' (0, 0, 1).

\end{itemize}

As output the FIS has the following variables: Intensity, Complexity,
Pattern and Mute/Unmute.

\begin{itemize}

\item \textit{Intensity.} Depends on Average Velocity, Intensity Shift
  and Time since sudden shift up/down. Controls the intensity parameter
  of the virtual drummer. U:(0, 127). Member functions shape:
  Triangular. Terms: 'Low' (0, 0, 25.4) 'Mid-Low' (0, 25.4, 50.8),
  'Middle' (25.4, 50.8, 76.2), 'Mid-High' (50.8, 76.2, 101.6), 'High'
  (76.2, 101.6, 127), 'Max' (101.6, 127, 127).

\item \textit{Complexity.} Depends on Full range density, Low range
  density, High range density, Pedal and Intensity. Controls the
  intensity parameter of the virtual drummer. U:(0, 127). Member
  functions shape: Triangular. Terms: 'Low' (0, 0, 25.4) 'Mid-Low' (0,
  25.4, 50.8), 'Middle' (25.4, 50.8, 76.2), 'Mid-High' (50.8, 76.2,
  101.6), 'High' (76.2, 101.6, 127), 'Max' (101.6, 127, 127).

\item \textit{Pattern.} Depends on Current Mode, Historic Mode, Historic
  Pattern, Historic Mute, Complexity, Intensity, Bar, Velocity Change,
  Hype, Time in bar, High range density, Low range density and Time
  since pedal. Controls the patterns played by the drummer. U:(0,14)
  Member functions shape: Singular. Terms: 'Intro To Chorus 1' (0),
  'Fill To Chorus 1' (1), 'Fill To Chorus 2' (2), 'Fill To Chorus 3'
  (3), 'Fill To Chorus 4' (4), 'Fill To Chorus 5' (5), 'Fill To Chorus
  6' (6), 'Fill To Chorus 7' (7), 'Chorus 1' (8), 'Fill 1' (9), 'Outro'
  (10), 'None' (11), 'No change' (12), 'Fill 4' (13), 'Fill 3' (14).

\item \textit{Mute.} Depends on High range density, Time since pedal,
  Intensity, Bar, Velocity Change and Historic Mute. Controls which
  parts of the kit should be muted. U:(0, 2). Member functions shape:
  Singular. Terms: 'None' (0), 'Kick' (1), 'Kick and Snare' (2).

\item \textit{Unmute.} Depends on Intensity, Bar, Velocity Change and
  Historic Mute. U:(0, 2). Member functions shape: Singular. Terms:
  'None' (0), 'Kick' (1), 'Kick and Snare' (2).

\end{itemize}

As for rule evaluation, we use standard Mamdani implication and
composition-based inference, and the Center-of-Area defuzzification
method~\cite{Driankov.book2001}.


%% file: develop.tex
\section{Development and Testing}
\label{develop.sec}

\subsection{System development}

The system is implemented using Python (version 3.6.8). For parsing the
MIDI messages received as input and for sending MIDI messages to the
virtual drummer, the Python library MIDO (version 1.2.9) is used. The
virtual drummer used is Strike 2 (version 2.0.7) developed by AIR Music
Technology. As input the system takes the output of a MIDI capable
piano, either being performed live and connected to the system via a
MIDI capable sound-card or prerecorded and saved as a MIDI file.

\subsection{Knowledge elicitation}

The project includes people from computer science, music performance,
audio engineering and philosophy.  This highly inter-disciplinary nature
requiree a careful process for the conceptual and practical development.
Throughout the project, participants have kept journals on their
thoughts, and various interaction means have been used --- discussions,
workshops, shared documents, examples of piano performance, and system
demos.  In the initial phases, piano recordings were analyzed by the
performer himself through a process of open coding, where different
features of the playing were identified and described; e.g. ``phrase
with high intensity'', ``build up in velocity'', etc.  These indications
then provided a basis for identifying the relevant musical parameters
and fuzzy rules in the AI system.

As it appears from the individual journals, the interaction has led to
cross fertilization and mutual enrichment of all the participants.  For
example, the need to describe music performance in logical terms led to
the development of a new analytical perspective on how, when and why
different styles are being chosen and used.  On the other hand, the
fuzzy models had to be enriched to meet the complexity of human musical
performance, e.g., to change the feeling of intensity in the music using
density of notes, change of notes registries, sustain pedal, or
dynamics.
Each representative of their field of science has also been interfacing
between the other two. For example, in discussion between music
performer and computer scientist the audio engineer has been able to
fill in with details on MIDI protocol and a possible different approach,
and in discussions between computer scientist and audio engineer about
technical aspects, the music performer has been able to expand the
discussion towards the above mentioned complexity in human musical
performance.
 
By building the design of the project on Strike as a auditive engine,
there could be more focus at the development of interaction between
human performer and the AI responsiveness and anticipation. Instead of
studying the basics of drum performance in terms of how and when to hit
the different part of the drumkit, the project could leave that to
Strike and focus on how to change the expressiveness of the virtual
drummer in accordance with the piano performance. More time has in this
way been put to interpret and interact to the piano (human) performance
instead of how to sound like a drummer.

\subsection{Testing}

The project was done from the start in a tight loop between the
musicians and the software developers.  To allow this, we have first
developed a simple but fully usable system, and then modified the system
and the model incrementally in collaboration with the musicians.  At the
time of this writing, the system has not been evaluated by an external
audience yet.  This will happen soon in three public concerts, two on
May 28 and one on June 12, 2019.


%% file: conclusions.tex
\section{Conclusions}
\label{conclusions.sec}

The presented case study is the first step in a more general and
ambitious study on anticipation and creative processes in anthrobotic
systems~\cite{crea.web}.  The work is at a preliminary stage, and the
treatment of anticipation is extremely simple and limited to a naive
prediction of a coming climax.  Despite this simplicity, the resulting
behavior of the virtual drummer was deemed sufficiently collaborative
and ``human like'' by the musicians in our initial tests.  The system
will be used for three live performances in the next month (June 2019),
and we will update this article to reflect the outcomes of this.

Anticipation needs a predictive model.  In the current system, this
model was elicited from the musicians and encoded in the form of fuzzy
predicates and fuzzy rules.  The use of a pure knowledge-based approach
in this initial phase allowed us to go through a modular and incremental
development in a continuous dialogue with the music experts.  Our next
step will be to integrate this approach with a data-driven approach,
e.g., by tuning, completing or adapting the rules from samples as done
in~\cite{FribergEtAl.acp2006}.  We also plan to investigate the use of
our framork to control different robot performers, e.g., a dancing
robot.


%% file: main.bbl
\begin{thebibliography}{10}

\bibitem{AdamyKempf.fss2003}
J{\"u}rgen Adamy and Roland Kempf.
\newblock Regularity and chaos in recurrent fuzzy systems.
\newblock {\em Fuzzy Sets and Systems}, 140(2):259--284, 2003.

\bibitem{Biles.leo2003}
John~A Biles.
\newblock Genjam in perspective: a tentative taxonomy for ga music and art
  systems.
\newblock {\em Leonardo}, 36(1):43--45, 2003.

\bibitem{Cope.book2005}
David Cope.
\newblock {\em Computer models of musical creativity}.
\newblock MIT Press Cambridge, 2005.

\bibitem{DeMirandaEtAl.rp2016}
L~{De Miranda}, M~Rovatsos, and S~Ramamoorthy.
\newblock We, {Anthrobot}: Learning from human forms of interaction and esprit
  de corps to develop more plural social robotics.
\newblock In {\em Proceedings of Robo-Philosophy: What social robots can and
  should do:}, pages 48--59, 2016.

\bibitem{crea.web}
Luis {De Miranda} and Alessandro Saffiotti.
\newblock Cross-disciplinary research on effectual anticipation.
\newblock \url{http://aass.oru.se/Agora/CREA}, 2019.

\bibitem{Driankov.book2001}
Dimiter Driankov.
\newblock A reminder on fuzzy logic.
\newblock In D~Driankov and A~Saffiotti, editors, {\em Fuzzy Logic Techniques
  for Autonomous Vehicle Navigation}, chapter~2. Springer, 2001.

\bibitem{FribergEtAl.acp2006}
Anders Friberg, Roberto Bresin, and Johan Sundberg.
\newblock Overview of the kth rule system for musical performance.
\newblock {\em Advances in Cognitive Psychology}, 2(2-3):145--161, 2006.

\bibitem{HillerIsaacson.jaes1958}
Lejaren~A Hiller~Jr and Leonard~M Isaacson.
\newblock Musical composition with a high-speed digital computer.
\newblock {\em Journal of the Audio Engineering Society}, 6(3):154--160, 1958.

\bibitem{LiuChuanKang.etci2017}
Chien-Hung Liu and Chuan-Kang Ting.
\newblock Computational intelligence in music composition: A survey.
\newblock {\em {IEEE} Transactions on Emerging Topics in Computational
  Intelligence}, 1(1):2--15, 2017.

\bibitem{Minsky.mmb1982}
Marvin Minsky.
\newblock Music, mind, and meaning.
\newblock {\em Computer Music Journal}, 3(4):28--44, 1982.

\bibitem{MorEtAl.acme2019}
Bhavya Mor, Sunita Garhwal, and Ajay Kumar.
\newblock A systematic literature review on computational musicology.
\newblock {\em Archives of Computational Methods in Engineering}, pages 1--15,
  2019.

\bibitem{Rosen.book2012}
Robert Rosen.
\newblock Anticipatory systems.
\newblock In {\em Anticipatory systems}, pages 313--370. Springer, 2012.

\bibitem{SimonSumner.mmm1968}
Herbert~A Simon and Richard~K Sumner.
\newblock Patterns in music (1968).
\newblock pages 83--110, 1993.
\newblock Reprinted.

\bibitem{TatarPasquier.jnmr2019}
K{\i}van{\c{c}} Tatar and Philippe Pasquier.
\newblock Musical agents: A typology and state of the art towards musical
  metacreation.
\newblock {\em Journal of New Music Research}, 48(1):56--105, 2019.

\bibitem{Thom.um2003}
Belinda Thom.
\newblock Interactive improvisational music companionship: A user-modeling
  approach.
\newblock {\em User Modeling and User-Adapted Interaction}, 13(1-2):133--177,
  2003.

\end{thebibliography}
